\documentclass[a4paper,fleqn]{cas-dc}
\usepackage[numbers, sort&compress]{natbib} 

\usepackage{threeparttable}
\usepackage{flushend}
\usepackage{graphicx}
\usepackage{subcaption}

\UseRawInputEncoding

\def\tsc#1{\csdef{#1}{\textsc{\lowercase{#1}}\xspace}}
\tsc{WGM}
\tsc{QE}

\begin{document}
\begin{sloppypar} 

\let\WriteBookmarks\relax
\def\floatpagepagefraction{1}
\def\textpagefraction{.001}
\let\printorcid\relax

\makeatletter
\renewcommand{\fnum@figure}{Fig. \thefigure.\@gobble}
\makeatother

\shorttitle{}    

\shortauthors{}  

\title [mode = title]{SaliencyI2PLoc: saliency-guided image-point cloud localization using contrastive learning}  


\tnotetext[1]{This research was supported by the National Natural Science Foundation Project (No.42130105, No. 42201477, No. 42101446){, Jiangxi Provincial Natural Science Foundation of (Grant No. 20242BAB21014), Open Research Fund Program of LIESMARS (Grant No. 23S02).}} 

\author[1]{Yuhao Li}

\ead{yhaoli@whu.edu.cn}
\credit{Conceptualization of this study, Methodology, Software, Validation, Writing – original draft}
\affiliation[1]{organization={State Key Laboratory of Information Engineering in Surveying, Mapping and Remote Sensing},
            addressline={Wuhan University}, 
            city={Wuhan},
            postcode={430079}, 
            country={China}}

\author[2]{Jianping Li}
\cormark[1]
\ead{jianping.li@ntu.edu.sg}
\credit{Conceptualization of this study, Methodology, Writing review, Funding acquisition}
\affiliation[2]{organization={School of Electrical and Electronic Engineering},
            addressline={Nanyang Technological University}, 
            country={Singapore}}

\author[1]{Zhen Dong}
\cormark[1]
\ead{dongzhenwhu@whu.edu.cn}
\credit{Conceptualization of this study, Methodology, Writing review, Project administration, Funding acquisition}

\author[3]{Yuan Wang}
\ead{wangyuanwhu@jxnu.edu.cn}
\credit{Methodology, Writing review}
\affiliation[3]{organization={School of Geography and Environment},
            addressline={Jiangxi Normal University}, 
            city={Nanchang},
            postcode={330022}, 
            country={China}}

\author[1]{Bisheng Yang}
\ead{bshyang@whu.edu.cn}
\credit{Conceptualization of this study, Project administration, Funding acquisition}

\cortext[1]{Corresponding author}



\begin{abstract}
Image to point cloud global localization is crucial for robot navigation in GNSS-denied environments and has become increasingly important for multi-robot map fusion and urban asset management. The modality gap between images and point clouds poses significant challenges for cross-modality fusion. Current cross-modality global localization solutions either require modality unification, which leads to information loss, or rely on engineered training schemes to encode multi-modality features, which often lack feature alignment and relation consistency. To address these limitations, we propose, \textbf{SaliencyI2PLoc}, a novel contrastive learning based architecture that fuses the saliency map into feature aggregation and maintains the feature relation consistency on multi-manifold spaces. To alleviate the pre-process of data mining, the contrastive learning framework is applied which efficiently achieves cross-modality feature mapping. The context saliency-guided local feature aggregation module is designed, which fully leverages the contribution of the stationary information in the scene generating a more representative global feature. Furthermore, to enhance the cross-modality feature alignment during contrastive learning, the consistency of relative relationships between samples in different manifold spaces is also taken into account. Experiments conducted on urban and highway scenario datasets demonstrate the effectiveness and robustness of our method. Specifically, our method achieves a Recall@1 of 78.92\% and a Recall{@20} of 97.59\% on the urban scenario evaluation dataset, showing an improvement of 37.35$\%$ and 18.07$\%$, compared to the baseline method. This demonstrates that our architecture efficiently fuses images and point clouds and represents a significant step forward in cross-modality global localization. The project page and code will be released at \url{https://whu-lyh.github.io/SaliencyI2PLoc/}.

\end{abstract}


\begin{highlights}
\item We introduce the \textbf{SaliencyI2PLoc}, a Dual-Transformer-based architecture for images to point clouds global localization tasks. 
\item The contrastive learning is leveraged to reduce the preprocessing of explicit data mining for sample pairs.
\item The saliency maps are fused into local patch feature aggregation module to create more representative global features.
\item The feature relation consistency is considered and integrated to ensure the consistent mapping of cross-modality features.

\end{highlights}



\maketitle

\section{Introduction}

The Global Navigation Satellite System (GNSS) is the most widespread localization tool, however, its effectiveness sharply declines in scenarios with GNSS signal obstruction, such as urban canyons~\citep{liMultiGNSSPPPINS2023,liMuCoGraphMultiscaleConstraint2023,yuanCooperativeLocalizationDisconnected2017}. Localization capabilities in GNSS-denied scenarios are indispensable for mobile robot navigation and autonomous driving~\citep{leeFailSafeMultiModalLocalization2022,zhanabatyrovaAutomaticMapUpdate2023}. Fusing diverse sensing data, such as the on-board cameras and pre-build LiDAR point cloud maps, has been considered an alternative for inferring positions relative to city-scale maps without relying on GNSS~\citep{miaoSurveyMonocularReLocalization2024a,yuBraininspiredMultimodalHybrid2023,huangMultimodalPolicyFusion2023,liHCTOOptimalityawareLiDAR2024}.

Compared to pre-built maps based on images, pre-built LiDAR point cloud maps offer rich geometric details, are less susceptible to adverse weather and lighting conditions, and pose lower privacy leakage risks. Consequently, they are commonly used as references for positioning. While equipping systems with expensive LiDAR can be cost-prohibitive, cameras provide a more versatile and practical solution for positioning on robots and other consumer-level platforms. Cross-modality fusion of images and point clouds strikes a balance between cost and performance, offering a more flexible and feasible approach to positioning~\citep{miaoSurveyMonocularReLocalization2024a}. However, the modality gaps present significant challenges to cross-modality fusion and localization, which remain in their infancy~\citep{zhaoAttentionEnhancedCrossmodalLocalization2023}.

\begin{figure}
\centering{
\includegraphics[scale=0.45]{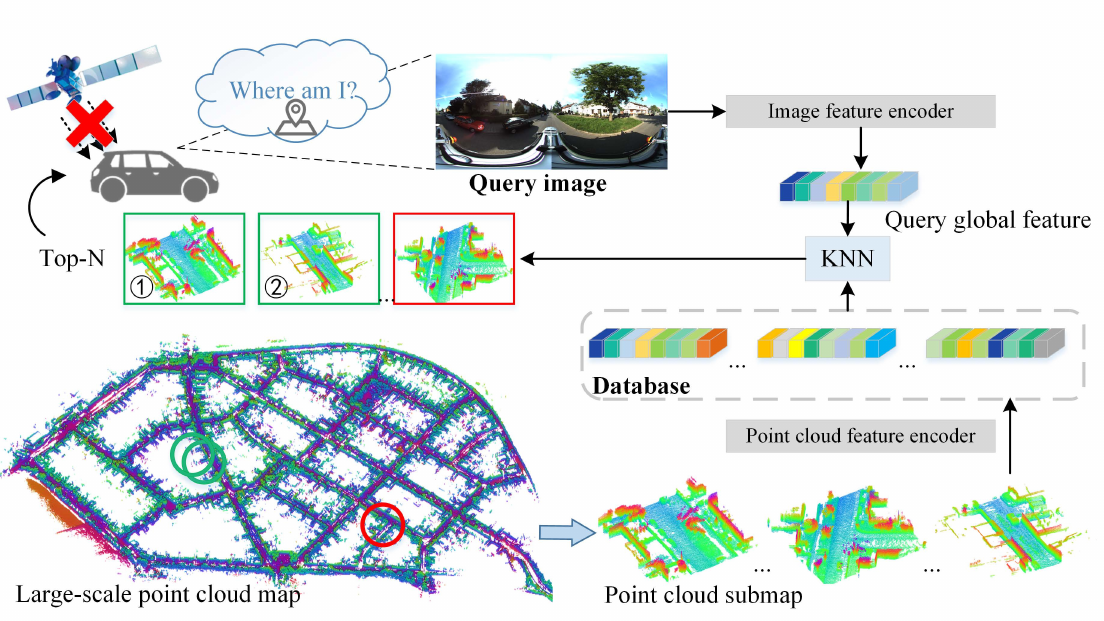}}
\caption {Overview of general cross-modality global localization. Given a point cloud map and a query image, the cross-modality localization task aims to retrieve the most closet or similar pre-built point cloud submaps.}
\label {fig_teaser}
\vspace{-0.6cm}
\end{figure}

This paper focuses on achieving the fusion of images and point clouds to enable {coarse} visual localization of a single image within a pre-built point cloud map, as illustrated in Fig.~\ref{fig_teaser}. Specifically, given a pre-built point cloud map and an input query image, the task is to identify the most similar point cloud submap within a specified distance tolerance. This task can be viewed as a retrieval task, where global features are computed for the input query image and then used to retrieve point cloud submaps from a database composed of global features generated from the point cloud submaps.

In this cross-modality localization task, the gap between modalities results in significant differences in feature descriptions. Intuitively, images capture the appearance and brightness of scenes in a structured manner in 2D space, while point clouds focus on capturing geometric information by discretely storing the 3D coordinates of the scene. However, common feature extractors used in the image domain cannot be directly applied to point clouds. {A big} challenge is to fuse data from multiple modalities into the same latent space. Existing solutions to eliminate modality gaps can be classified into two categories: modality transformation-based solution and dual-tower model-based solution.

\textbf{Modality transformation-based solution} converts different types of data into a unified representation before employing a common feature extractor for feature fusion~\citep{bernreiterSphericalMultiModalPlace2021, yinI3dLocImagetorangeCrossdomain2021, zhengI2PRecRecognizingImages2023, leeLC2LiDARCameraLoop2023, xieModaLinkUnifyingModalities2024}. Specifically, this approach either involves lifting images to 3D space using predicted depths or projecting point clouds onto 2D images. However, modality transformation relies on specific imaging patterns, which can introduce additional errors during conversion, such as instability in depth prediction or dimensionality loss from 3D to 2D.

\textbf{Dual-tower model-based solution} uses modality-specific feature encoders to separately encode images and point clouds, aiming to integrate multi-modality data into the same high-dimensional latent space~\citep{cattaneoGlobalVisualLocalization2020, zhaoAttentionEnhancedCrossmodalLocalization2023, liVXPVoxelCrossPixelLargescale2024}. These methods are typically based on Siamese network architectures and begin by constructing cross-modality positive and negative sample pairs. Through phased training strategies, they aim to fuse multi-modality feature representations and ensure feature alignment. However, current methods often overlook the contribution of salient objects in the scene to global feature descriptors, resulting in insufficient scene representation.
Additionally, current global localization approaches use metric learning based on positive and negative sample pairs, which often require the explicit construction of \textit{triplets} or \textit{quadruplets} from existing datasets. This involves mapping multi-modality data to the same latent space through complex data mining strategies. Inadequate construction of positive and negative pairs can severely impact feature representation. Moreover, as datasets expand, the need to reconstruct positive and negative sample pairs arises, which can be cumbersome and lacks versatility.

To mitigate errors arising from modality conversion, this paper proposes a novel cross-modality image-point cloud localization method employing a dual-tower structure and contrastive learning framework. Our approach employs a Siamese architecture to directly process 2D images and 3D point clouds, and utilizes a self-attention-based feature extractor to extract local patch features from each modality. We leverage context attention between these local patch features to generate powerful global features. In contrast to supervised learning methods that require the pre-construction of \textit{triplets}, our approach achieves feature alignment of multi-modality data through contrastive learning. Furthermore, to ensure consistency in feature representation across different samples and modalities within multi-manifold spaces, we introduce a multi-manifold space feature relation consistency loss as the supervision. The main contributions are as follows:

\begin{itemize}
\item We propose \textbf{SaliencyI2PLoc}, a Dual-Transformer-based architecture for image-point cloud localization, leveraging contrastive learning for cross-modality fusion, thereby avoiding the rigid preprocessing of data mining.
\item To enhance the representation of stationary information in the scene, we integrate the saliency map into local patch feature aggregation, thereby producing more representative global features.
\item To improve feature alignment between images and point clouds, we integrate cross-modality feature relation consistency into contrastive learning, thereby ensuring consistent mapping between images and point clouds.
\end{itemize}

The remainder of this paper is organized as follows. Section~\ref{sec_rw}  presents a review of global localization and multi-modality representation. Section~\ref{sec_method} elaborates on the proposed method in detail. The description of the dataset, implementation details, qualitative and quantitative results, and ablation studies are given in Section~\ref{sec_exp}. Finally, {the} conclusions and future work are summarized in Section~\ref{sec_con}.

\section{Related work}
\label{sec_rw}

This section offers a detailed review of global localization tasks based on images and point clouds, covering both unimodal and multimodal fused global localization approaches. Additionally, we {provide} a brief overview of feature learning frameworks designed specifically for multi-modality data.

\subsection{Unimodality global localization}
Unimodality global localization focuses on acquiring global feature descriptors and enhancing retrieval performance. Traditional methods such as Vector of Locally Aggregated Descriptors (VLAD)~\citep{arandjelovicAllVLAD2013}, aggregate hand-crafted local feature descriptors to accomplish image retrieval. With the outstanding performance of neural networks in feature representation, NetVLAD~\citep{arandjelovicNetVLADCNNArchitecture2016} pioneered the differentiable of VLAD, enabling end-to-end training of feature aggregation modules. Additionally, lightweight aggregation modules~\citep{radenovicFineTuningCNNImage2019} have shown remarkable effectiveness. To leverage local information fully, methods such as Patch-NetVLAD~\citep{hauslerPatchNetVLADMultiScaleFusion2021}, TransVPR~\citep{wangTransVPRTransformerBasedPlace2022}, and Hybrid-CNN-Vit~\citep{wangHybridCNNTransformerFeatures2023} aggregate local features of multi-scales to generate more powerful global features. Spatial topology~\citep{sheImagePatchMatchingGraphBased2023} and semantic information~\citep{xueEfficientLargescaleLocalization2022, pirasInformationFusionContent2017} of scenes are also considered to achieve better scene representation. Furthermore, the introduction of re-ranking strategies~\citep{zhuR2formerUnifiedRetrieval2023,kannanPlaceFormerTransformerbasedVisual2024,shaoGlobalFeaturesAre2023} in the retrieval process greatly enhances the precision. With the development of large visual foundation models, fine-tuning or transfer learning on pre-trained models leads to improved performance~\citep{izquierdoOptimalTransportAggregation2024, luSeamlessAdaptationPretrained2024, keethaAnyLocUniversalVisual2024,wangFreeRegImagetoPointCloud2023}.

Due to the susceptibility to lighting, weather, and season changes, images may exhibit significant variations, whereas point clouds can provide more accurate geometry. Therefore, point clouds-based global localization methods have gradually emerged. Handcrafted feature descriptors such as SSC~\citep{kimScanContextEgocentric2018} perform well in non-learning methods. PointNetVLAD~\citep{uyPointNetVLADDeepPoint2018}, by combining NetVLAD structures with PointNet~\citep{charlesPointNetDeepLearning2017}, achieves the first end-to-end global localization based on point clouds. Furthermore, incorporating local geometric information of point clouds enriches the expression capability of global features~\citep{zhouNDTTransformerLargeScale3D2021, liuLPDNet3DPoint2019,zhangPCAN3DAttention2019}. Methods like PPTNet~\citep{huiPyramidPointCloud2021} and HiTPR~\citep{houHiTPRHierarchicalTransformer2022} aggregate multi-scale local features to cover more contextual information, and the local features further boost the retrieval percision~\citep{vidanapathiranaSpectralGeometricVerification2023,zhangRankPointRetrievalRerankingPoint2023}. 
Sparse 3D convolution-based backbone achieves a trade-off in both efficiency and performance~\citep{komorowskiMinkLoc3DPointCloud2021}. Meanwhile, schemes based on multi-view projection of point clouds~\citep{maCVTNetCrossViewTransformer2023,zhaoSphereVLADAttentionBasedSignalEnhanced2023} achieve robustness in rotation invariance in point cloud-based global localization. Currently, frameworks for point cloud global localization are diverse, with no unified representation model yet established.

\subsection{Multi-modality global localization}
Multi-modality global localization aims to leverage the characteristics of data from different modalities, the emphasis lies in feature fusion and reducing modality gaps. These approaches often employ dual-tower models, where PIC-Net~\citep{luPICNetPointCloud2020} utilize LPD-Net~\citep{liuLPDNet3DPoint2019} and ResNet~\citep{heDeepResidualLearning2016} to extract features from point clouds and images respectively, and then fuse modality-specific features using multilayer perceptron. Compared to shallow fusion, feature fusion at higher depths has achieved good results~\citep{panCORALColoredStructural2021,laiAdaFusionVisualLiDARFusion2022,yuMMDFMultiModalDeep2022,liuMFFPRPointCloud2022,zhouLCPRMultiScaleAttentionBased2023,panCameraLiDARFusionLatent2024,yueCrossFusionPoint2024}. DistillVPR~\citep{wangDistilVPRCrossModalKnowledge2024} integrates the knowledge distillation to transfer the scene representation capability of 3D feature extractors to image feature extractors, thereby improving the precision of global localization. {A pioneering cross-modality localization method proposed by Cattaneo et al.~\citep{cattaneoGlobalVisualLocalization2020}, encodes images and point clouds respectively, and utilizes carefully designed training strategies to achieve better feature mapping.} Zhao et al.~\cite{zhaoAttentionEnhancedCrossmodalLocalization2023} enhance the feature representation capability by adding channel attention in different modality encoders. VXP~\citep{liVXPVoxelCrossPixelLargescale2024} {proposes a two-stage training strategy to improve the joint representation performance of cross-modal features. In the first stage, Transformer and 3D sparse convolution are used to encode image and point cloud features, respectively, and the voxelized features are mapped onto the image feature map to optimize the local features between the pixel and voxel feature maps. In the second stage, the generalized feature pooling method GeM~\citep{radenovicFineTuningCNNImage2019} is used to obtain the global feature vector, and model parameters are optimized based on a triplet loss function}. 
However, aligning features across different modalities remains challenging. Transforming different modalities into the same embedding space, such as Bird's Eye View (BEV) space~\citep{zhengI2PRecRecognizingImages2023,xuLeveragingBEVRepresentation2023}, spherical space~\citep{yinI3dLocImagetorangeCrossdomain2021,bernreiterSphericalMultiModalPlace2021,leeLC2LiDARCameraLoop2023}, or range images~\citep{xieModaLinkUnifyingModalities2024}, is advantageous for reducing modality differences. However, modal transformation requires data extrinsic as prior, and projection may rely on special scanning patterns of sensors, which restricts its applications.

\subsection{Multi-modality feature representation}
Research focusing on how to encode multi-modality data in a shared embedding space has become a hotspot~\citep{radfordLearningTransferableVisual2021,ermolovHyperbolicVisionTransformers2022,pengHyperbolicDeepNeural2022,afhamCrossPointSelfSupervisedCrossModal2022a}. The Transformer blocks~\citep{vaswaniAttentionAllYou2017a} serve as the core unit, demonstrating excellent performance in {multiple cross-modality tasks~\citep{zhangSelfSupervisedImageDenoising2023,zhouHighDynamicRange2023}. The self-attention mechanism inside Transformer could capture the long-range context and significantly improve the performance on many scene understanding tasks~\citep{fangCrossmodalityAttentiveFeature2022, liuHidformerHierarchicalDualtower2024}}. Also, the contrastive learning training strategies are gradually becoming a basic paradigm~\citep{heMomentumContrastUnsupervised2020,arunMultimodalEarthObservation2022}. In global localization tasks, the current paradigms generally employ \textit{triplet} or \textit{quadruplet} to provide supervision~\citep{schroffFaceNetUnifiedEmbedding2015,arandjelovicNetVLADCNNArchitecture2016}, with efficient data mining being a top priority. However, the number and quality of negatives have a large influence on the feature representation~\citep{wiesmannKPPRExploitingMomentum2023}. Contrastive learning, the semi-supervised strategies based on multi-modality data pairs, can avoid the need for complex data mining strategies~\citep{bertonDeepVisualGeolocalization2022,hongDecoupledandCoupledNetworksSelfSupervised2023}, making them a more suitable choice for cross-modality global localization. LIP-Loc~\citep{shubodhLIPLocLiDARImage2024} adopts a contrastive learning manner by projecting point clouds onto range images and aligning features between images and range images proxy to achieve cross-modality localization. Our method directly handles the point cloud, avoiding unnecessary loss of dimensional information.

\section{SaliencyI2PLoc}
\label{sec_method}

\begin{figure*}
\includegraphics[scale=0.9]{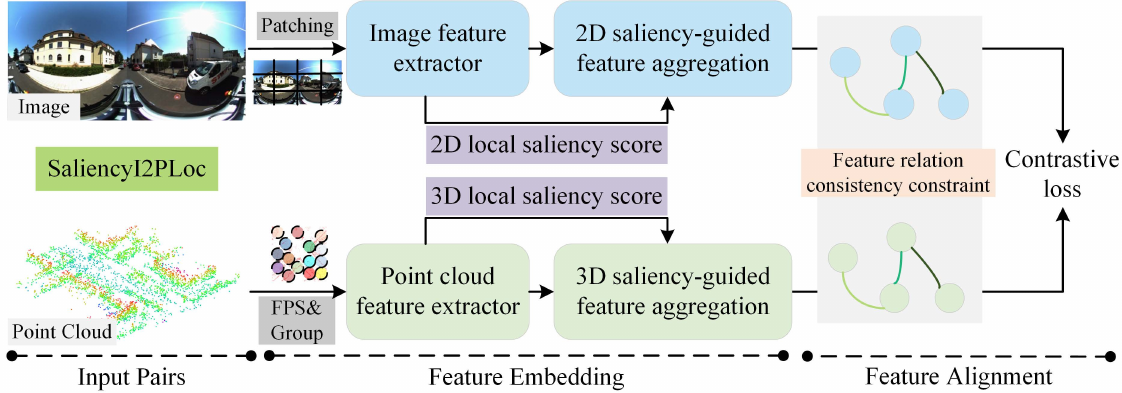}
\caption {The architecture of SaliencyI2PLoc. {SaliencyI2PLoc encodes the input image-point cloud pairs into a high-dimensional feature embedding space using a feature encoder (ViT for images, mini-PointNet combined with Transformer for point clouds) and feature aggregator (saliency map boosted NetVLAD layer). It then achieves feature fusion and alignment through the contrastive learning loss function that incorporates cross-modal feature relationship consistency constraints.}}
\label{fig_methodology}
\vspace{-0.5cm}
\end{figure*}

For the cross-modality fusion and localization between images and point clouds, the core of the proposed \textbf{SaliencyI2PLoc} lies in feature encoding and alignment, adopting a dual-tower architecture in the contrastive learning manner as illustrated in Fig. ~\ref{fig_methodology}.

It assigns images and point clouds into tokens and then maps the tokens into high-dimensional space based on Transformer blocks (Section~\ref{sec_fe}). The saliency score obtained in the feature extractor is leveraged and fused into the NetVLAD module, achieving adaptive feature weighting (Section~\ref{sec_agg}). Meanwhile, a feature relation consistency constraint is designed between samples in multi-manifold to provide supervision for better feature alignment. The definitions of contrastive loss and the feature relation consistency loss will be presented in Section~\ref{sec_loss}.

\subsection{Modality-specific feature extraction}
\label{sec_fe}

\subsubsection{Image feature extraction}

SaliencyI2PLoc converts raw images into tokens and utilizes a standard Vision Transformer (ViT)~\citep{dosovitskiyImageWorth16x162023} to extract local features from these tokens. Then the local features are aggregated as the global feature. Specifically, given a query image $I \in \mathbb{R}^{H \times W \times C}$, $H, W, C$ denote the height, width, and channel numbers of the image. For the input image, we do not change the resolution, and the raw image is converted into $N_{2d} = H \cdot W / P^2$ patches. $P$ is the side length of a square patch in the image. Through a learnable linear projection layer, these flattened patches are mapped into visual tokens $\mathbf{X} = \{x_1, x_2, \dots, x_{N_{2d}} \} \in \mathbb{R}^{N_{2d} \times D
_{2d}}$ as the input of the Transformer blocks, where $D_{2d}$ is the feature dimension of 2D patch embeddings. We set the $P$ to 16, and stack 12 Transformer blocks in the image branch. We also use the default class token and add learnable positional embedding to each token. The tokens will pass through three projection matrices to generate $\mathbf{Q}, \mathbf{K}, \mathbf{V} \in \mathbb{R}^{(N+1) \times D_{2d}}$. The self-attention module is formulated in the following manner:
\vspace{-0.4cm}
\begin{equation}
\text{Self-attention}(\mathbf{Q}, \mathbf{K}, \mathbf{V}) 
=\text{Softmax} \Big(\frac{\mathbf{Q} {\mathbf{K}}^{T}}{\sqrt{D_{2d}}}\Big) \mathbf{V}.
\vspace{0.5em}
\label{fig_attention}
\end{equation}

The scaled dot product $\mathbf{Attn} = \text{Softmax} \frac{\mathbf{Q} {\mathbf{K}}^{T}}{\sqrt{D_{2d}}} $ indicates the correlation between patches. The higher the attention value, the more related the patch is. The classical structure of the $l$-th Transformer block could be represented as:
\begin{align}
\mathbf{X}'_l &= \mathbf{X}_l + \text{Self-attention}_l(\mathbf{X}_l), \\
\mathbf{X}_{l+1} &= \mathbf{X}'_l + \text{FFN}_l(\mathbf{X}'_l),
\end{align}
where $\text{FFN}$ represents the feed-forward network consisting of successive linear layers and layer normalization operation.

\subsubsection{Point cloud feature extraction}

As point clouds encompass extensive areas and rich geometric context, it is crucial to incorporate this information for effective long-range feature encoding. SaliencyI2PLoc utilizes a lightweight PointNet~\citep{charlesPointNetDeepLearning2017} as a tokenizer to encode local point cloud patches into tokens. Transformers are introduced to capture global relationships among these local patch features. This approach ensures that Transformer blocks endow different modalities with a consistent perception pattern across a large receptive field, while maintaining a detailed representation of the scene.

Specifically, given an input point cloud $\mathbf{P} \in \mathbb{R}^{N'_{3d} \times 3}$, where $N'_{3d}$ is the number of the point cloud, we first sample $N_{3d}$ points randomly by farthest point sampling (FPS). Then, the $k$ nearest points around each $N_{3d}$ as center points are fetched via the k-nearest neighbor (KNN) algorithm. We convert raw point cloud $\mathbf{P}$ into $N_{3d}$ patches, which are then normalized by subtracting corresponding center points. The 3D patches are fed into the 3D tokenizer that is built based on PointNet to project the raw data into 3D patch embeddings $\mathbf{F}_{3d} \in \mathbb{R}^{N_{3d} \times D_{3d}}$, where $D_{3d}$ is the feature dimension of 3D patch embeddings. 

\begin{figure}
\centering{
\includegraphics[scale=0.48]{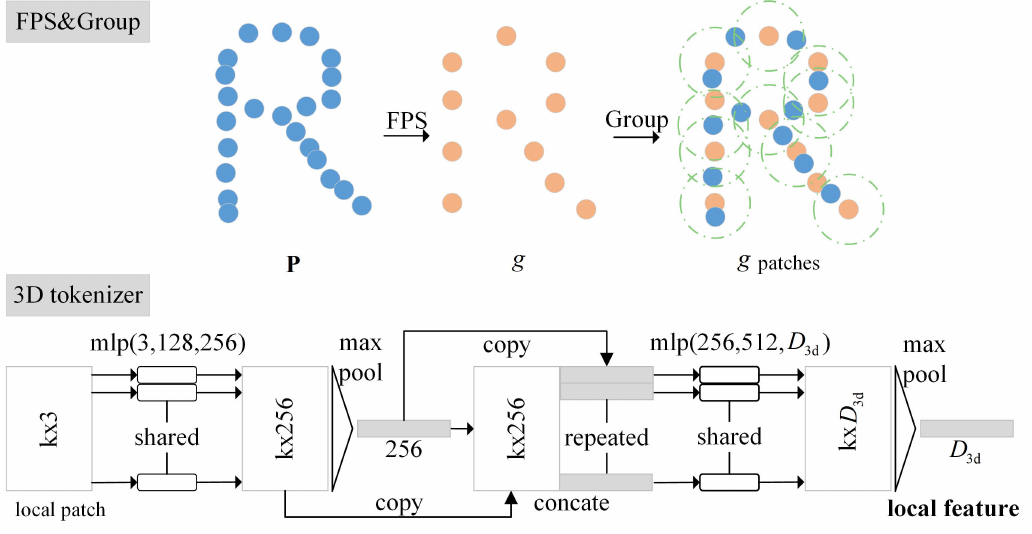}}
\caption {Visualization of farthest point sampling (FPS) and grouping process, and the architecture of 3d tokenizer. Numbers in the bracket are layer sizes for the Multi-layer perceptron (MLP). Batchnorm is used for all layers with ReLU. 3D tokenizer takes in local patch point clouds and returns a $D_{3d}$-dimensional feature.}
\label{fig_tokenizer3D}
\vspace{-0.5cm}
\end{figure}

The 3D tokenizer is visualized in Fig.~\ref{fig_tokenizer3D}, and we simplify the structure of PointNet and utilize the shared MLP to extract the local feature for each patch. The patch embeddings $\mathbf{F}_{3d}$ are treated as point cloud tokens $\mathbf{Y} = \{y_1, y_2, \dots, y_{N_{3d}} \} \in \mathbb{R}^{N_{3d} \times D
_{3d}}$. We also apply 12 Transformer blocks to boost the interaction between local patch features. We discard the class token and adopt the learnable position embedding for each depth of the Transformer blocks. The local features are calculated through the same self-attention structures and the transformer blocks as:
\begin{align}
\mathbf{Y}'_l &= \mathbf{Y}_l + \text{Self-attention}_l(\mathbf{Y}_l), \\
\mathbf{Y}_{l+1} &= \mathbf{Y}'_l + \text{FFN}_l(\mathbf{Y}'_l).
\end{align}

\subsection{Saliency-guided feature aggregation}
\label{sec_agg}

\begin{figure}
\centering{
\includegraphics[scale=0.55]{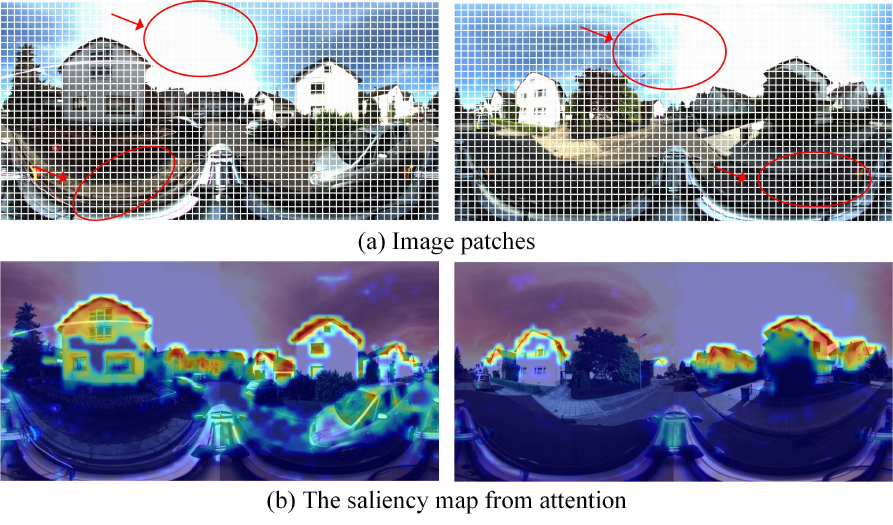}}
\vspace{-0.5cm}
\caption {Illustration of image patchifying and the salient attention map. (a) The patches of the input images, where the invalid areas which provide less information are highlighted by red ellipses. (b) The activation maps from the last Transformer blocks of the visual encoder are back-projected into input images, where the lighter inflects higher attention scores.}
\label {fig_netvlad_attention}
\vspace{-0.5cm}
\end{figure}

Images contain invalid areas such as sky and ground as shown in Fig. ~\ref{fig_netvlad_attention} (a), that could not provide representative information and the contribution of the local features in these areas is limited. Similarly, point clouds face a similar situation due to occlusion and view direction. However, for localization tasks, the overall scenario layout and the stationary objects in the scene should be focused during representation. The local features extracted from these saliency areas should contribute more while representing the scene. Thus a self-attention mechanism from the Transformer-based feature encoder is leveraged. The saliency map is generated during local feature interaction indicating the relative relations. We then fuse the saliency map into the feature aggregation module to achieve adaptively local feature weighting, enhancing the representativeness of the global feature.

NetVLAD~\citep{arandjelovicNetVLADCNNArchitecture2016} is the common feature aggregation module to capture the statistical information from residuals between local features and clusters. Formally, given $N$ $D$-dimension local features $\mathbf{F} = \{f_1, f_2, \dots, f_N \} \in \mathbb{R}^{N \times D}$ and predefined $K$ cluster centers $\mathbf{C} = \{c_1, c_2, \dots, c_K \} \in \mathbb{R}^{K \times D}$, the vanilla version of VLAD feature $\mathbf{V} \in \mathbb{R}^{D \times K}$ could be obtained by:
\vspace{-0.2cm}
\begin{align}
V(j, k) &= \sum_{i=1}^{N} a_k(\mathbf{f}_i) (f_i(j) - c_k(j)), \\
a_k(\mathbf{f}_i) &= \frac{e^{\mathbf{w}_k^T \mathbf{f}_i + b_k}}{\sum_{k'}{e^{\mathbf{w}_{k'}^T \mathbf{f}_i + b_{k'}}}},
\end{align}
where $f_i(j)$ and $c_k(j)$ are the $j$th dimension of $i$th local feature and $k$th cluster center. $a_k(\mathbf{f}_i)$ is the soft assignment function in which the $\mathbf{w}_k, b_k$ are learnable parameters.  

\begin{figure}
\centering{
\includegraphics[scale=0.48]{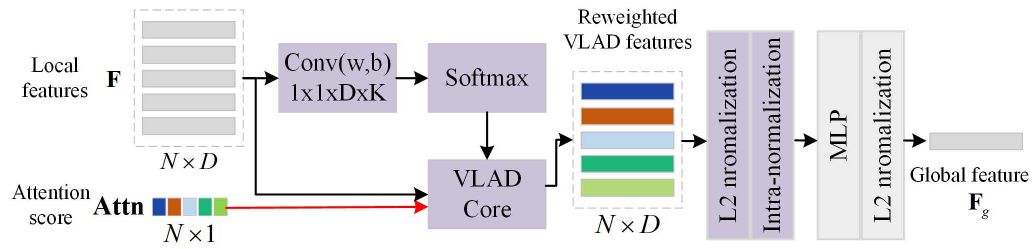}}
\caption {The pipeline of saliency-guided NetVLAD layer. The red arrow indicates the position where the saliency score is applied, while the purple blocks are the vanilla NetVLAD layer.}
\label {fig_netvlad}
\vspace{-0.5cm}
\end{figure}

As shown in Fig. ~\ref{fig_netvlad_attention} (b), we project the activation map of the last Transformer blocks of our visual encoder to the input image. It could be seen that the activation map demonstrates the saliency region of the scene, especially the stationary areas. The local features from patches with higher attention should contribute more to the description. We propose to assign the attention score $\mathbf{Attn} \in \mathbb{R}^N$ from the last Transformer blocks to the NetVLAD layer, as shown in Fig. ~\ref{fig_netvlad}, which forces the VLAD feature to focus on task-relevant regions. Specifically, we multiply attention maps by local patch features, and the refined NetVLAD could be formulated as:
\begin{equation}
V(j, k) = \sum_{i=1}^{N} \mathbf{Attn} \cdot a_k(\mathbf{f}_i) (f_i(j) - c_k(j)).
\end{equation}

The VLAD feature is flattened as a $(D \times K)$-dimension feature. We then reduce the feature dimension $\mathbf{V} \in \mathbb{R}^{(D \times K)}$ to the desired embedding size $\mathbf{F}^g \in \mathbb{R}^{D_g}$ by a linear projection layer. Because we have adopt a similar structure based on the Transformer for feature extraction in both image and point cloud modalities, the attention from the final layer of Transformer blocks in different modalities can be extracted and utilized for the NetVLAD structure mentioned above. Specifically, for images, the $D$ equals to $D_{2d}$, while for point clouds, it is set to $D_{3d}$. We set the $K, D_g$ to 64 and 256 for both modalities, respectively. Finally, the global feature for image and point cloud is calculated as:
\begin{align}
\mathbf{F}^g_{2d} &= \text{Attn-NetVLAD}(\mathbf{F}_{2d}) = \text{Attn-NetVLAD}(\mathbf{X}), \\
\mathbf{F}^g_{3d} &= \text{Attn-NetVLAD}(\mathbf{F}_{3d}) = \text{Attn-NetVLAD}(\mathbf{Y}).
\end{align}

\subsection{Loss functions}
\label{sec_loss}

To project the features from images and point clouds in the shared space and achieve cross-modality alignment, we firstly adopt the flexible {and widely used} contrastive learning loss, InfoNCE loss~\citep{oordRepresentationLearningContrastive2019, hongDecoupledandCoupledNetworksSelfSupervised2023}, as the supervision for end-to-end training. InfoNCE forces the embeddings of positive image-point cloud samples to be close and those of negative samples to be distant. However, the feature relations of unimodality samples are not consistent across multi-modalities. Thus, we propose to supervise the distribution of the multi-modality features through multi-manifold constraints and enhance the feature relation consistency across multi-modalities.


Formally, assume that we obtain a $B$-sample mini-batch of image-point cloud pairs, after passing through the SaliencyI2PLoc, we obtain global features $\mathbf{F}^g_{2d} = \{\mathcal{F}_1^{2d}, \mathcal{F}_2^{2d}, \dots, \mathcal{F}_B^{2d} \} \in \mathbb{R}^{B \times D_g}$ from images and $\mathbf{F}^g_{3d} = \{\mathcal{F}_1^{3d}, \mathcal{F}_2^{3d}, \dots, \mathcal{F}_B^{3d} \} \in \mathbb{R}^{B \times D_g}$ from point clouds, respectively. The $\mathcal{F}_i^{2d}, 1 \leq i \leq B$ and $\mathbf{F}^g_{3d}$ are considered as the query and the keys respectively from the perspective of dictionary $\textit{look-up}$ task. The InfoNCE loss is formulated as:
\begin{equation}
\label{eq_infonce}
\mathcal{L}_{c}(\mathcal{F}_i^{2d}, \mathbf{F}^g_{3d}) 
= -\log \frac{\exp\left(\mathcal{F}_i^{2d} \cdot \mathcal{F}_i^{3d} / \tau\right)}
{\sum_{j=1}^{B} \exp\left(\mathcal{F}_i^{2d} \cdot \mathcal{F}_j^{3d} / \tau\right)},
\end{equation}
where $\mathcal{F}_i^{3d}$ denotes the cross-modality positive sample corresponding to $\mathcal{F}_i^{2d}$, the $\mathcal{F}_j^{3d}$ that $j \neq i$ samples in the same mini-batch are automatically treated as negatives. The dot product represents the cosine similarity between samples, and $\tau$ is the temperature hyper-parameter.

\begin{figure}
\centering{
\includegraphics[scale=1.0]{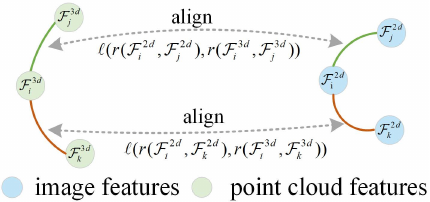}}
\vspace{-0.2cm}
\caption {The demonstration of cross-modality feature relation consistency.}
\label {fig_feature_align}
\vspace{-0.5cm}
\end{figure}

\textbf{Feature consistency constraints}. Once the features across multi-modality are well projected in the same shared embedding space, the relative feature consistency between different samples should be identical. We measure the relation difference of two samples across modalities. Specifically, Given two image-point cloud pairs in mini-batch $B$, we obtain the corresponding feature $\mathcal{F}_i^{2d}, \mathcal{F}_j^{2d}$ from images $I_i, I_j$ and $\mathcal{F}_i^{3d}, \mathcal{F}_j^{3d}$ from point clouds $\mathbf{P}_i, \mathbf{P}_j$, respectively. As shown in Fig.~\ref{fig_feature_align}, {we apply the distance formulas in Euclidean spaces to measure the feature relationship between two features. In Euclidean space, the relation function is represented as Euclidean distance $D_{euc}(\cdot, \cdot)$ between two features.} And we consider maintaining the feature relation consistency across different modalities. The feature relation consistency loss is formulated as:
\begin{align}
\label{eq_2d3d_s}
\mathcal{L}_{rel} = \sum_{i,j\in B} \ell (r(\mathcal{F}_i^{2d}, \mathcal{F}_j^{2d}), r(\mathcal{F}_i^{3d}, \mathcal{F}_j^{3d})),
\end{align}
where $r(\cdot, \cdot)$ is the relative relation function to compute embedding distances, {$\ell$ represents the MSE (Mean Square Error) loss.}

\textbf{Multi-manifold constraint}. The dot product between two different modality features confines the features in the Euclidean space, however, the distribution of the multi-modality feature is sophisticated. We also explore fusing the feature relation consistency in the Hyperbolic space~\citep{nickelPoincareEmbeddingsLearning2017,ermolovHyperbolicVisionTransformers2022}. The Hyperbolic space has constant negative curvature and a different distance metric. We use the common $n$ dimensions \textit{Poincar\'e ball} $(\mathbb{D}_c^n, \mathbf{g}^{\mathbb{D}})$ model of Hyperbolic space with the curvature parameter $c$ (the actual curvature value is $-c^2$). The $\mathbb{D}_c^n$ is defined as $\mathbb{D}_c^n= \{ \mathbf{q} \in \mathbb{R}^n \colon c\| \mathbf{q} \|^2 < 1, c \geq 0\}$ and Riemannian metric $\mathbf{g}^{\mathbb{D}} = \lambda_c^2 \mathbf{g}^E$, where $ \lambda_c = \frac{2}{1-c\| \mathbf{q} \|^2}$ is the conformal factor and $\mathbf{g}^E = \mathbf{I}_n$ is Euclidean metric identity tensor. Please refer to ~\citep{ermolovHyperbolicVisionTransformers2022} for more details.

Because of that the Hyperbolic spaces are not vector spaces, for a pair $\mathbf{q}, \mathbf{p} \in \mathbb{D}^n_c$, the \textit{Mobius addition} operation $\oplus_c$ in Hyperbolic space is formulated as:
\begin{equation}
\mathbf{q} \oplus_c \mathbf{p} = \frac{(1+2c\langle \mathbf{q}, \mathbf{p} \rangle + c\|\mathbf{p}\|^2) \mathbf{q}+ (1-c\|\mathbf{q}\|^2)\mathbf{p}}{1+2c\langle \mathbf{q}, \mathbf{p} \rangle + c^2 \|\mathbf{q}\|^2 \|\mathbf{p}\|^2}.
\end{equation}

The Hyperbolic distance between $\mathbf{q}, \mathbf{p} \in \mathbb{D}^n_c$ is given as:
\begin{equation}
\label{eq_hdist}
D_{hyp}(\mathbf{q},\mathbf{p}) = \frac{2}{\sqrt{c}} \mathrm{arctanh}(\sqrt{c}\|-\mathbf{q} \oplus_c \mathbf{p}\|).
\end{equation}

From Euclidean space to Hyperbolic space, the \textit{exponential} mapping is required. For some fixed base point $\mathbf{u} \in \mathbb{D}^n_c$, usually set to $\mathbf{0}$, the \textit{exponential} mapping $\exp_\mathbf{q}^c \colon \mathbb{R}^n \to \mathbb{D}_c^n$ is defined as:
\begin{equation}
\label{eq_exp}
\exp_\mathbf{u}^c(\mathbf{v}) = \mathbf{u} \oplus_ c \bigg(\tanh \bigg(\sqrt{c} \frac{\lambda_\mathbf{u}^c \|\mathbf{v}\|}{2} \bigg) \frac{\mathbf{v}}{\sqrt{c}\|\mathbf{v}\|}\bigg).
\end{equation}

{In the Hyperbolic space the feature relation is expressed as the Hyperbolic distance $D_{hyp}$.} 
Considering the multi-manifold constraints on Euclidean space and Hyperbolic space, the feature relation consistency loss defined is explicitly set as:
\begin{equation}
\begin{aligned}
\label{eq_2d3d_c_hyp}
\mathcal{L}^{fuse}_{rel}\ &= \lambda \mathcal{L}_{rel}^{euc} + \beta \mathcal{L}_{rel}^{hyp} \\
&= \lambda \sum_{i,j\in B} \ell (D_{euc}(\mathcal{F}_i^{2d}, \mathcal{F}_j^{2d}), D_{euc}(\mathcal{F}_i^{3d}, \mathcal{F}_j^{3d})) \\
&+ \beta \sum_{i,j\in B} \ell (D_{hyp}(\mathcal{F}_i^{2d}, \mathcal{F}_j^{2d}), D_{hyp}(\mathcal{F}_i^{3d}, \mathcal{F}_j^{3d})),
\end{aligned}
\end{equation}
where $D_{euc}$ represents the dot product, the $\lambda$ and $\beta$ are weight hyper-parameters to balance the specific losses. In the implementation, we set the $\lambda = 1$ and the $\beta$ to 2.0. The features are mapped into the Hyperbolic space by \textit{exponential} operation while calculating the hyperbolic distance.

The total loss function consists of InfoNCE loss and feature relation consistency loss and is formulated as:
\begin{align}
\label{eq_total_loss}
\mathcal{L} = \mathcal{L}_{c} + \mathcal{L}^{fuse}_{rel}.
\end{align}

\vspace{-0.5cm}
\section{Experiments}
\label{sec_exp}

\subsection{Dataset and evaluation metrics}
\subsubsection{Dataset generation}

\begin{figure*}[t]
\centering{
\includegraphics[scale=0.92]{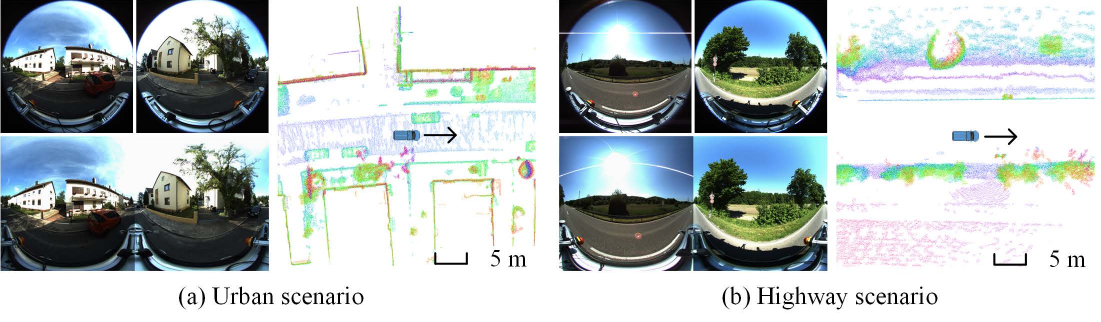}}
\vspace{-0.2cm}
\caption {Image-point cloud pair dataset for cross-modality global localization. This dataset covers different scenarios, such as urban which contains buildings and vehicles, and the highway scene which contains feature-less structures. (a) displays the panoramic image at the left-down and the bird's view of the point cloud submap of the urban scene, while (b) visualizes the highway scenario. The point cloud submaps are rendered by relative height. Black arrows indicate the moving direction of vehicles.}
\label {fig_dataset_kitti360}
\vspace{-0.5cm}
\end{figure*}

There are no publicly available datasets for cross-modality global localization tasks, we leverage the KITTI-360~\citep{liaoKITTI360NovelDataset2023}, {KITTI~\citep{geigerAreWeReady2012}} dataset and build an image-point cloud dataset for verifying the proposed method. The KITTI-360 dataset comprises 11 sequences with a total length of 73.7 km. There is little overlap between different sequences, but the majority of sequences contain revisits and loop closures. The dataset contains data from various scenes, mainly including urban environments with rich architectural structures, as well as open highways that mainly contain context-less objects. Additionally, the scenes include moving objects such as vehicles, pedestrians, trees, and feature-less textures. We separated the dataset into the training set and evaluation set, the latter contains two types of scenarios: urban and highway scenarios, as displayed in Fig. ~\ref{fig_dataset_kitti360}. 

{KITTI odometry dataset is widely used in autonomous driving and contains 21 sequences and 11 sequences with ground truth poses. The data covers various scenarios including urban areas, suburban areas, and highways, with dynamic interference from pedestrians, vehicles, and other moving objects in the scenes. Following the VXP~\citep{liVXPVoxelCrossPixelLargescale2024}, we train the model on sequences 03, 04, 05, 06, 07, 08, 09, 10 and evaluate the cross-modality localization on 00 sequence.}

Considering the significant information gap between the limited field of view of frame-based images and the coverage range of point clouds, our method was evaluated using surrounding view images with a 360-degree field of view. The KITTI-360 dataset contains multi-beam point clouds, perspective stereo images, two fisheye images, and GPS/IMU to obtain precious locations. The fisheye camera was with a 185-degree field of view and was installed on both sides of the vehicle, thus enabling omnidirectional scene perception. We used the FFMPEG\footnote{https://ffmpeg.org/} tool to stitch the fisheye images into an omnidirectional image, as shown in Fig.~\ref{fig_dataset_kitti360}. The size of the panoramic image was resized as $512 \times 1024$. The point cloud frames were merged into submaps based on the poses, and the length of the submap was $40 \times 40m$. We used the RANSAC method to remove the majority of ground points from the point cloud. The image and point cloud data were both accurately timestamped and calibrated, ensuring consistency in their coordinate systems. The dataset generation process will be released with the source code together.

We selected 1501 pairs from urban scenario sequences, and 1010 pairs from highway scenario sequences, for model evaluation. Additionally, we separated 3447 and 842 image-point cloud pairs from urban and highway scenarios respectively to evaluate the generalization of our method. These pairs do not exist in the training set, and are officially treated as the evaluation dataset for other scene understanding-related tasks. Considering the limitation of GPU memory and training time, we conducted the ablation experiments on the query image-point cloud pairs whose dataset size was almost 1/4 of the whole dataset. Following~\citep{zhaoAttentionEnhancedCrossmodalLocalization2023, warburgMapillaryStreetLevelSequences2020}, we selected an image as a query every 3 meters, with the remaining position serving as the database. The details of the dataset partition are listed in the Tab.~\ref{tab_dataset_kitti360}.

{The KITTI dataset does not include panoramic images, we directly evaluate the algorithm using perspective images. To ensure a fair comparison, we did not generate point cloud submaps, but instead performed cross-modality localization based on the raw LiDAR frames. The query and database data partitioning method follows the same approach as VXP~\citep{liVXPVoxelCrossPixelLargescale2024}.}

\begin{table}[!t]
\renewcommand{\arraystretch}{1.3}	
\caption {The dataset details of both urban and highway scenarios.}
\label{tab_dataset_kitti360}
\resizebox{0.48\textwidth}{!}{
\begin{tabular}{lllllllll}
\toprule
&\multicolumn{3}{c}{Urban Scenario}  & \multicolumn{3}{c}{Highway Scenario}  \\ \cline{2-7}
& Query & Database & Total & Query & Database & Total
\\\midrule 
Train           & 19256     & 58423     & 77679     & -         & -         & -      \\ 
Evaluation      & 332       & 1169      & 1501      & 404       & 606       & 1010   \\ 
Further Test    & 902       & 2545      & 3447      & 376       & 466       & 842    \\
\bottomrule
\end{tabular}%
}
\vspace{-0.5cm}
\end{table}

\subsubsection{Evaluation metrics}

We set the Euclidean distance threshold $\eta$ between the query and candidate point cloud submap (frame) as 20$m$, if the distance is smaller than $\eta$, the localization for the query is judged as successful. The Recall@N and max F1-score are used to evaluate the performance of cross-modality localization. In addition, the precision-recall curve is plotted to better visualize the performance.

\textbf{Recall@N}. $N \in \{1, 5, 10, 15, 20\}$ is the number of the candidate point cloud submaps to be retrieved. The Recall@N is the fraction of the Top-N score over the total query set and could be presented as:
\begin{align}
\label{recall@N}
\text{Top-N}(I_q, N) &= \left \{ 
\begin{array}{cc} 
    1, & if \sum_n^N \text{match}(I_q,I_{gt}(n)) \geq 1 \\
    0, & \text{otherwise}
\end{array} \right. , \\
\text{Recall@N} &= \frac{\sum_{j=1}^{N_{query}} \text{Top-N}(I_q(j), N) } {N_{query}}
\end{align}
where the $I_q$ is the query image, the $I_{gt}$ is the corresponding ground truth submaps, and $\text{match}(\cdot, \cdot)$ function returns \text{1}, if the localization for the query is successful.

\textbf{F1-score}. The F1 score is calculated based on precision and recall. We calculated the F1-score only at Top-1 and the F1-score could be defined as the following equation:
\begin{equation}
F_1= 2 \times \frac{P \times R}{P+R}=\frac{2\times TP}{2\times TP+FP+FN}.
\end{equation}

{According to the~\cite{houEvaluationObjectProposals2018}, the cosine similarity between whole query images and the corresponding Top-1 retrieved point cloud submaps are first calculated and stored in descending order. Then, the ranked feature similarities are sequentially used as thresholds $\xi$. For a given query image and retrieved Top-1 candidate point cloud submap, a true positive (TP) is defined if the feature similarity between the two is greater than $\xi$ and the Euclidean distance between retrieved Top-1 and the query image is smaller than $\eta$. If the Euclidean distance is larger than $\eta$, it is a false positive (FP). If the feature similarity between the two is less than $\xi$, and the Euclidean distance is smaller than $\eta$, it is defined as a false negative (FN), otherwise, it is defined as a true negative (TN).}

\subsection{Implementation details}

Given an input omnidirectional image size of $512 \times 1024$ at the KITTI-360 dataset, we obtained 2048 image patches. We initialized the ViT model using the weights pre-trained on the ImageNet dataset, and we froze the Transformer blocks except for the last 4 blocks following ~\citep{izquierdoOptimalTransportAggregation2024}. The position tokens are interpolated to adapt the current image resolution. We randomly sampled 3072 points from the input point clouds. For each sampled point, we searched 32 nearest points to construct 3D local patches. We used CUDA to accelerate the FPS and reduce the computational complexity. The features of each point cloud patch are encoded to 1024 and then reduced to 384, which is also the feature dimension of tokens input into the ViT. Due to GPU memory constraints, we applied a regular size of the transformer, specifically, we stacked 12 transformer blocks with 6 heads for the image feature extractor and 3 heads for the point cloud feature extractor. In the feature aggregation module, we adopted a uniform setting of 64 cluster centers, and the final global feature dimension is set to 256. The temperature parameter used to calculate the InfoNCE loss is set to the default value of 0.07.

The model undergoes end-to-end training on NVIDIA RTX 4090 GPU with 24 GB, running for 100 epochs with a batch size of 4. {We also use the gradient accumulation strategy to simulate a larger batch size}. The accumulate step is set to 64. During network training, we used the SGD optimizer with a learning rate of 0.0001 for the image feature extractor, 0.00001 for the point cloud feature extractor, and 0.0005 for the aggregator. The weight decay is set to 0.001. Except for the baseline scheme, we used the cosine learning rate scheduler, with a warm-up epoch of 3.

{The image resolution in the KITTI dataset is $375 \times 1242$, we resized the images to $368 \times 1232$ to ensure they are divisible by 16, allowing them to be fed into the ViT model. At the same time, we adjusted the batch size to 6, while keeping other hyperparameters unchanged.}

During network training, we simultaneously input image-point cloud pairs. However, during practical cross-modality global localization, only unimodality of data is required. Initially, we employed the point cloud encoder to encode the point cloud submap, saving the global feature descriptor onto disk as a database. For the given query image, only the forward pass of the visual encoder is necessary to obtain the visual global feature.

\subsection{Baseline methods}

A similar work to ours is AE-Spherical~\citep{zhaoAttentionEnhancedCrossmodalLocalization2023}, which utilized the spherical convolution to extract the local feature from panoramic images and applied the PointNetVLAD~\citep{uyPointNetVLADDeepPoint2018} for point clouds. The channel attention is used to enhance the representativeness of global features. They did not provide testable model weights\footnote{https://github.com/Zhaozhpe/AE-CrossModal} and the datasets are not available. Considering the miner difference during the dataset generation, we adopted the experimental setup and dataset setting from their paper and retrained the model as a baseline. We had trained many times, however, the results listed in the paper could not be reproduced following their training settings. The comparable results are listed in the Tab. ~\ref{tab_recall_baseline}.

We compare our SaliencyI2PLoc with the LIP-Loc~\citep{shubodhLIPLocLiDARImage2024}, which aligns the multi-modality data by modality transformation. Specifically, they converted the point clouds frames into range images based on the scan pattern of the laser scanner. They resized the input image to $224 \times 224$ to accommodate the default resolution settings of the ViT. Then the visual transformer-based feature encoder and linear projection layer are utilized to generate the global feature representations for images and range image proxy respectively. We also retrained their model\footnote{https://github.com/Shubodh/lidar-image-pretrain-VPR} by their default parameters with our datasets, and the results are presented in Tab.~\ref{tab_recall_baseline}.

{The model proposed by Cattaneo et al.~\citep{cattaneoGlobalVisualLocalization2020} and VXP~\citep{liVXPVoxelCrossPixelLargescale2024} are described in detail in the related work and used for comparison. However, the baseline model~\citep{cattaneoGlobalVisualLocalization2020} is not publicly available and we used the implementation provided by VXP. We also retrained both model~\citep{cattaneoGlobalVisualLocalization2020} and VXP\footnote{https://github.com/yunjinli/vxp} with their default parameters on the KITTI dataset.}

\subsection{Performance analysis}

\begin{table*}[t]
\centering
\caption{Recall value on the  KITTI-360 evaluation dataset. We highlight the \textbf{best} results.}
\label{tab_recall_baseline}
\begin{threeparttable}
\renewcommand{\arraystretch}{1.3}	
\resizebox{0.96\textwidth}{!}{
\begin{tabular}{llllllllll}
\toprule
\multirow{2}{*}{Method types}                   & \multirow{2}{*}{Method}                                                         & \multicolumn{4}{c}{Urban Scenario}                                         & \multicolumn{4}{c}{Highway Scenario}                                     \\ \cline{3-10}
                                                &                                                                                 & R1$(\%)\uparrow$& R5 $(\%)\uparrow$& R20 $(\%)\uparrow$& Max F1 $\uparrow$ & R1$(\%)\uparrow$& R5$(\%)\uparrow$& R20 $(\%)\uparrow$& Max F1$\uparrow$ \\ \midrule 
\multirow{3}{*}{Dual-tower model-based}& AE-Spherical\tnote{1} ~\citep{zhaoAttentionEnhancedCrossmodalLocalization2023}  & 41.57           & 60.54            & 79.52             & 0.59              & 10.64           & 22.28           & 45.30             & 0.31             \\ 
                                                & AE-Spherical*\tnote{2} ~\citep{zhaoAttentionEnhancedCrossmodalLocalization2023}   & 46.23           & 66.04            & 75.47             & ***               & 10.00           & 19.00           & 39.00             & ***          \\    
                                                & {SCNN-Contrast\tnote{3}}   & {60.24}           & {69.58}            & {85.84}             & {0.75}               & {22.77}           & {37.13}           & {60.40}             & {0.37}              \\ \hline
\multirow{1}{*}{Modality transformation-based}      & LIP-Loc ~\citep{shubodhLIPLocLiDARImage2024}                                     & 64.46           & 79.52            & 91.27             & 0.79              & \textbf{37.87}  & \textbf{58.42}  & \textbf{79.46}    & \textbf{0.55}    \\ \hline
\multirow{2}{*}{Dual-tower model-based}& \textbf{SaliencyI2PLoc}-1/4\tnote{4} (ours)                                     & 69.28           & 81.63            & 93.98             & 0.82              & 26.73           & 42.82           & 65.10             & 0.43             \\
                                                & \textbf{SaliencyI2PLoc (ours)}                                                  & \textbf{78.92}  & \textbf{86.75}   & \textbf{97.59}    & \textbf{0.88}     & 30.94           & 49.26           & 75.99             & 0.47             \\ \bottomrule
\end{tabular}}
\begin{tablenotes}
\footnotesize
\item[1] {The re-trained baseline model by ourselves will be used in further evaluation.}
\item[2] Some values are not clearly marked in the original paper, which are estimated from the figure.
\item[3] {The feature encoder is applied based on baseline AE-Spherical~\citep{zhaoAttentionEnhancedCrossmodalLocalization2023} but is trained using the contrastive learning loss function InfoNCE.}
\item[4] These models are trained on the 1/4 of whole image-point cloud pair datasets.
\end{tablenotes}
\end{threeparttable}
\vspace{-0.5cm}
\end{table*}

\subsubsection{Quantitative results}

\begin{table}[!t]
\centering
\renewcommand{\arraystretch}{1.3}	
\caption{{Recall value on the KITTI dataset. We highlight the \textbf{best} results.}}
\label{tab_recall_kitti}
\resizebox{0.48\textwidth}{!}{
\begin{tabular}{lllllllll}
\toprule
\multirow{2}{*}{Method}                                                & \multicolumn{4}{c}{KITTI 00 sequence}                   \\ \cline{2-5}
                                                                       & R1$(\%)\uparrow$ & R5 $(\%)\uparrow$ & R20 $(\%)\uparrow$ & Max F1 $\uparrow$ \\ \midrule 
AE-Spherical~\citep{zhaoAttentionEnhancedCrossmodalLocalization2023}   & 24.18            & 42.53             & 68.48              & 0.39                           \\ 
Cattaneo et al.~\citep{cattaneoGlobalVisualLocalization2020}           & 22.32            & 27.43             & 44.79              & 0.27                           \\ 
VXP~\citep{liVXPVoxelCrossPixelLargescale2024}                         & \textbf{41.01}   & 45.02             & 63.00              & 0.40                  \\ 
\textbf{SaliencyI2PLoc (ours)}                                         & 31.78            & \textbf{47.23}    & \textbf{72.59}     & \textbf{0.49}         \\ \bottomrule
\end{tabular}
}
\vspace{-0.5cm}
\end{table}

\begin{figure}[t]
\centering{
\includegraphics[scale=0.5]{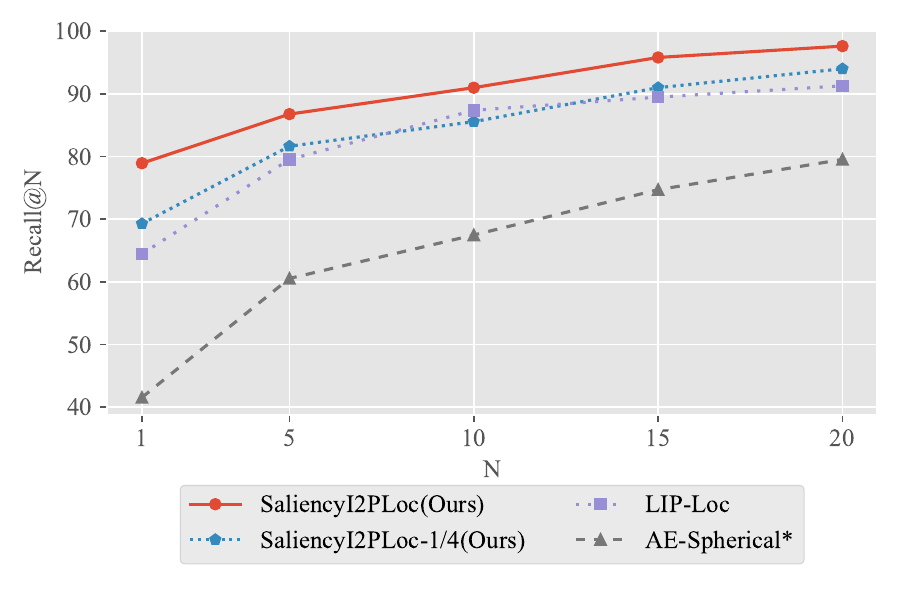}}
\vspace{-0.2cm}
\caption {The significant margin of recall compared to the baseline method {on KITTI-360 urban evaluation datasets.}}
\vspace{-0.5cm}
\label {fig_topn}
\end{figure}

As shown in Tab.~\ref{tab_recall_baseline}, our method demonstrates significant performance improvements compared to the baseline AE-Spherical. In evaluation dataset of urban scenario, our retrained baseline model achieves a Recall@1 of 41.57$\%$ and a Recall@20 of 79.52$\%$, while our SaliencyI2PLoc achieves 78.92$\%$ and 97.59$\%$ respectively. Compared to the AE-Spherical, our approach shows an improvement of 37.35$\%$ and 18.07$\%$. The Top-N curve presented in Fig.~\ref{fig_topn} clearly demonstrates the superiority of our method, as the recall values at different Top-N thresholds significantly surpass those of the baseline method. Compared to the modality transformation solution LIP-Loc~\citep{shubodhLIPLocLiDARImage2024}, our method also achieves improvements of 14.46$\%$ and 6.32$\%$ on the urban scenario datasets.

Under the data mining paradigm, the query set is treated as the anchor and the positive and negative sample pairs are collected from the database. Our method is built based on the contrastive learning framework which explicitly avoids positive and negative pair construction. We also trained the model using image-point cloud pairs from the query set, where the data volume is approximately a quarter of the total training set. Our method achieves a Recall@1 of 69.27$\%$ and a Recall@20 of 93.98$\%$ on the urban scenario evaluation dataset, significantly surpassing the baseline method AE-Spherical by margins of 27.70$\%$ and 14.46$\%$. These improvements attribute to the effective cross-modality fusion via the feature representation capabilities of the proposed architecture. Moreover, this demonstrates the scalability of our approach. With the expansion of new data, we believe our method can further enhance retrieval performance without engineered reconstruction of \textit{triplets}. {To further verify the superiority of the contrastive learning framework, we integrated the feature encoder from the baseline~\citep{zhaoAttentionEnhancedCrossmodalLocalization2023} into the architecture of our proposed method and trained it using the InfoNCE contrastive learning loss function. The quantitative evaluation results, as shown in Tab.~\ref{tab_recall_baseline}, demonstrate that the feature representation framework based on contrastive learning significantly outperforms the triplet-based method, achieving better multi-modality data feature fusion.}

\begin{figure}[t]
\centering{
\includegraphics[scale=0.5]{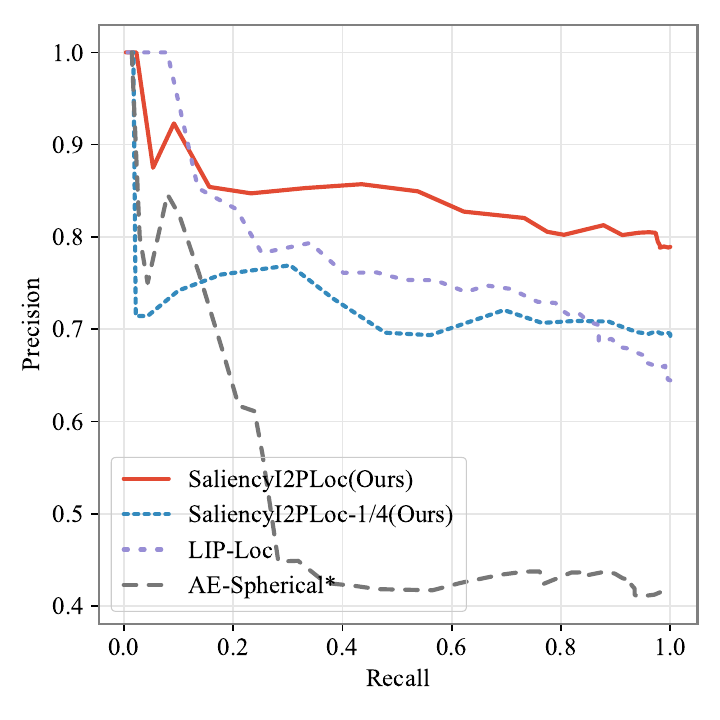}}
\vspace{-0.2cm}
\caption {{The precision-recall (PR) curve on KITTI-360 urban evaluation datasets.}}
\vspace{-0.6cm}
\label {fig_pr}
\end{figure}

We also present evaluation results on highway scenario to assess the possibility of global localization in open scenario without relying on GNSS. In highway scenario, the absence of significant landmarks presents a major challenge for feature extraction from 3D point clouds, leading to the failure of feature-based localization methodology. The flatten structure-less areas contribute insufficiently to the scene representation. The ability of describing scene appearance makes the image projection-based method more effective. Despite a decrease in performance in highway scenario, our method still outpaces the baseline AE-Spherical by margins of 20.30$\%$ and 30.69$\%$. It is worth to note that, for practical usage, the serviceability of GNSS at open areas is affordable, where fusing the multi-modalities could be a complementary. 

{As shown in the Tab.~\ref{tab_recall_kitti}, while VXP achieves the best Recall@1, our method is superior in terms of overall performance across a range of recall and F1 metrics. Our method shows an improvement of 9.59$\%$ in Recall@20 compared to VXP. It is worth note that the point clouds in the KITTI dataset are raw point cloud frames, so their density is lower compared to the point cloud submaps used in KITTI-360. However, the results on the KITTI dataset demonstrate that our method exhibits a certain level of robustness to point cloud density. }

\subsubsection{Qualitative results}
\begin{figure*}[!p]
\centering{
\includegraphics[scale=0.775]{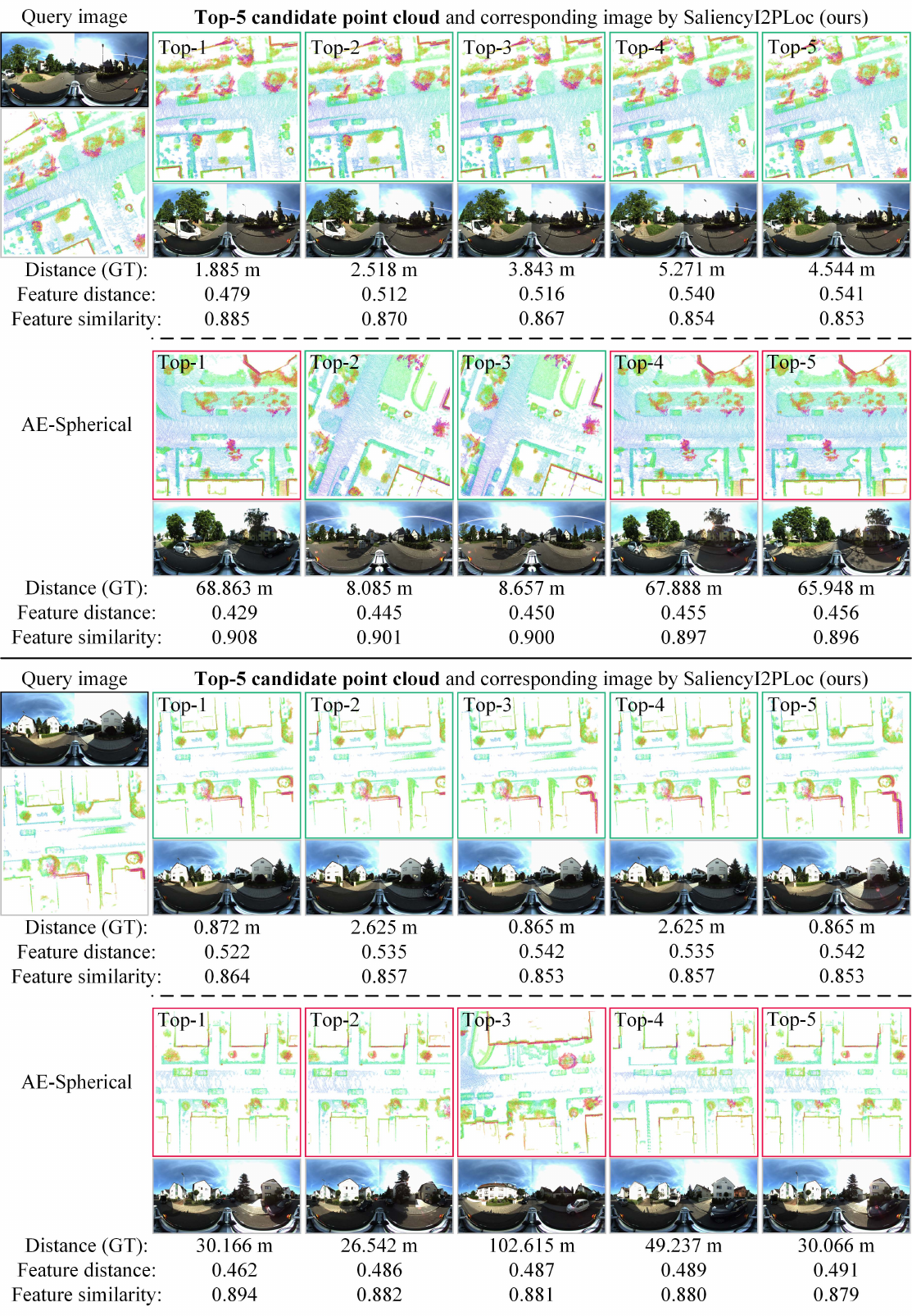}}
\caption {Visualization of the retrieval results on the KITTI-360 urban scenario evaluation dataset. The green boxes indicate the corrected results, while the red represents the failed ones. The images of each fetched point cloud are displayed for better visualization.}
\label{fig_retrieval_results}
\vspace{-0.5cm}
\end{figure*}

As shown in Fig.~\ref{fig_retrieval_results}, we visualize the cross-modality localization results, with images of Top-5 serving as auxiliary and not participating in the localization. Red boundaries indicate incorrect results, while green indicates the correct results. The point clouds are rendered by the relative height from the bird eye view. We also label the corresponding feature similarity scores and the ground truth Euclidean distances. It is apparent that our method consistently achieves effective localization results compared to the AE-Spherical, demonstrating its ability to fully utilize reliable objects within the scenes to generate more descriptive global descriptors. The feature distance indicates that the baseline model cannot correctly differentiate between scenes that are similar but different, while our method could effectively locate the correct places. 

\begin{figure}[!t]
\centering
\subfloat[AE-Spherical]{\includegraphics[width=0.235\textwidth] {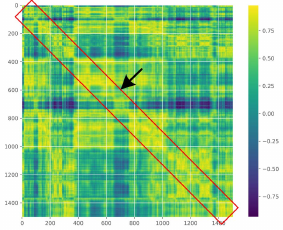}
\label{fig_similarity_compare-a}}
\hfil
\subfloat[SaliencyI2PLoc (ours)]{\includegraphics[width=0.235\textwidth] {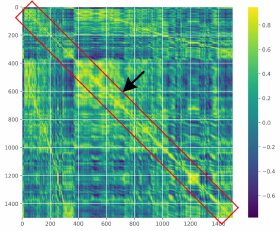}
\label{fig_similarity_compare-b}}
\caption {Visualization of the cosine similarity between image and point clouds of KITTI-360 the evaluation dataset. The color gradient from green to yellow in the figure indicates a gradual increase in similarity, where red boxes highlight the bright areas.}
\label{fig_similarity_compare}
\end{figure}

The cosine similarity between images and point clouds on the evaluation dataset is shown in Fig.~\ref{fig_similarity_compare}. The horizontal axis represents the image features of the evaluation set, while the vertical axis represents the corresponding point cloud features. The color gradient from green to yellow in the figure indicates a gradual increase in similarity. There is a prominent bright area along the diagonal direction as shown in Fig.~\ref{fig_similarity_compare-b}, which is highlighted by red boxes. The highlighted part in the figure indicates that our method can effectively map data from different modalities to the same latent space, achieving better feature fusion and alignment. 

\begin{figure}[!t]
\centering
\subfloat[AE-Spherical]{\includegraphics[width=0.24\textwidth] {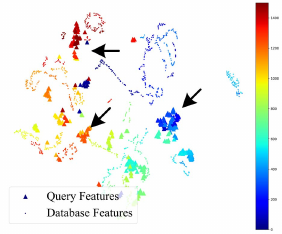}\label{fig_tsne_compare-a}}
\hfil
\subfloat[SaliencyI2PLoc (ours)]{\includegraphics[width=0.24\textwidth] {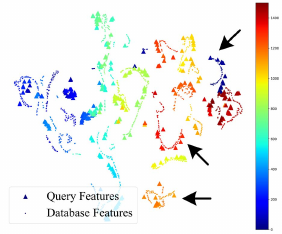}\label{fig_tsne_compare-b}}
\caption {Visualization of feature descriptors {on Euclidean space} using t-SNE~\citep{maatenVisualizingDataUsing2008}. The data index is rendered from blue to red in ascending order. $\blacktriangle$ and $\cdot$ represent the feature from the query and database respectively.}
\label{fig_tsne_compare}
\end{figure}

\begin{figure}[!t]
\centering
\subfloat[AE-Spherical]{\includegraphics[width=0.24\textwidth] {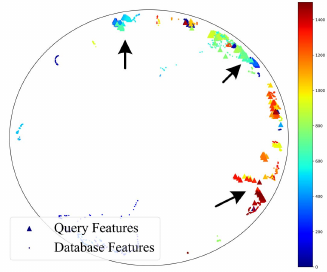}\label{fig_umap_compare-a}}
\hfil
\subfloat[SaliencyI2PLoc (ours)]{\includegraphics[width=0.24\textwidth] {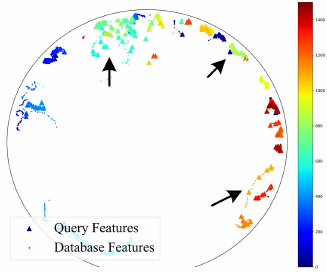}\label{fig_umap_compare-b}}
\caption {Visualization of feature descriptors {on Hyperbolic space} using UMAP. The data index is rendered from blue to red in ascending order. $\blacktriangle$ and $\cdot$ represent the feature from the query and database respectively.}
\label{fig_umap_compare}
\end{figure}

In addition, we employ t-SNE~\citep{maatenVisualizingDataUsing2008} {and UMAP\footnote{https://umap-learn.readthedocs.io/en/latest/embedding\_space.html\\\#bonus-embedding-in-hyperbolic-space}} to visualize the features of query and database on Euclidean space and Hyperbolic space. The whole image-point cloud pairs are time sequential arranged {at the KITTI-360 dataset}, and the queries were sampled at intervals of 3 meters. Thus, the features of queries and database data should be contiguous in terms of their IDs, resulting in similar rendered colors. Note that the index of the image and point cloud pair is only used for visualization. The position of the query feature is displayed by $\blacktriangle$, and the point cloud features of the database are rendered by $\cdot$. The black arrows highlight the crowded areas that result from the lack of distinctiveness of global features generated by the baseline AE-Spherical in Fig.~\ref{fig_tsne_compare-a} {and Fig.~\ref{fig_umap_compare-a}}. It is clearly shown in Fig.~\ref{fig_tsne_compare-b} {and Fig.~\ref{fig_umap_compare-b}} that our method effectively integrates features across modalities and makes features more distinctive.

\subsubsection{Generalization ability}
We further evaluate the generalization ability of our method by selecting samples that had never appeared in the training dataset. As shown in Tab.~\ref{tab_recall_generalization}, our method still achieves a decent recall performance. Compared to the baseline AE-Spherical, we maintain a significant performance advantage.

\begin{table*}[!t]
\begin{threeparttable}
\centering
\renewcommand{\arraystretch}{1.3}	
\caption{Recall value on the further dataset for generalization test. We highlight the \textbf{best} results.}
\label{tab_recall_generalization}
\begin{tabular}{lllllllll}
\toprule
\multirow{2}{*}{Method}              & \multicolumn{4}{c}{Urban Scenario}                                            & \multicolumn{4}{c}{Highway Scenario}                                             \\ \cline{2-9}
                                     & R1$(\%)\uparrow$ & R5 $(\%)\uparrow$ & R20 $(\%)\uparrow$ & Max F1 $\uparrow$ & R1$(\%)\uparrow$ & R5$(\%)\uparrow$  & R20 $(\%)\uparrow$    & Max F1 $\uparrow$ \\ \midrule 
AE-Spherical\tnote{1} ~\citep{zhaoAttentionEnhancedCrossmodalLocalization2023}  & 37.14            & 60.42             & 82.59              & 0.54              & 4.79             & 20.74             & 43.09               & 0.09              \\ 
\textbf{SaliencyI2PLoc (ours)}                                                  & \textbf{61.86}   & \textbf{81.15}    & \textbf{95.90}     & \textbf{0.77}     & \textbf{22.34}   & \textbf{50.27}    & \textbf{72.07}       & \textbf{0.37}     \\ \bottomrule
\end{tabular}%
\begin{tablenotes}
\footnotesize
\item[1] {The re-trained baseline model by ourselves.}
\vspace{-0.5cm}
\end{tablenotes}
\end{threeparttable}
\end{table*}

\subsection{Ablation analysis}
In this section, we primarily investigate the impact of the saliency map on the process of local feature aggregation, and the feature relation consistency will be discussed about cross-modality feature representation. Besides, we conducted an ablation study about the selection on number of 3D tokens.

\subsubsection{The efficiency of saliency-guided NetVLAD}
\begin{figure}[!t]
\centering{
\includegraphics[scale=0.89]{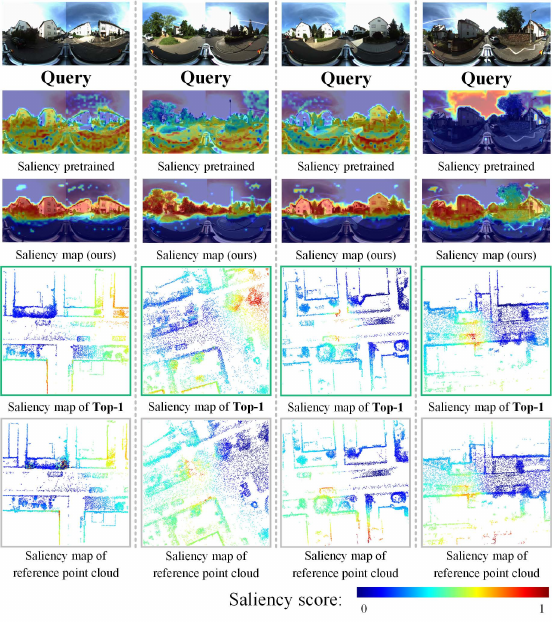}}
\caption {The visualization of saliency maps of the query images. During training, we notice that the saliency map shifts to the scene layout and stationary buildings.}
\label {fig_saliency_maps_changes}
\vspace{-0.5cm}
\end{figure}

As shown in Tab.~\ref{tab_recall_ablation}, utilizing a saliency-guided feature aggregation module effectively enhances the performance of our model. Pretrained weights on the ImageNet dataset comprehensively focus on whole semantic features within the scene, including the front-end semantic objects, as presented in Fig. ~\ref{fig_saliency_maps_changes}. Throughout the training process, we continuously fine-tuned them to be adaptable to cross-modality localization tasks. Fig. ~\ref{fig_saliency_maps_changes} illustrates that our feature encoders can capture features of saliency objects in the scene, such as buildings, road signs, and stationary vehicles. These objects contribute more to the scenario description and establish a more distinctive global feature representation. Therefore, fusing saliency maps into the feature aggregation module is an effective operation, further enhancing the scene representation capability.

\begin{figure*}[!t]
\centering{
\includegraphics[scale=0.775]{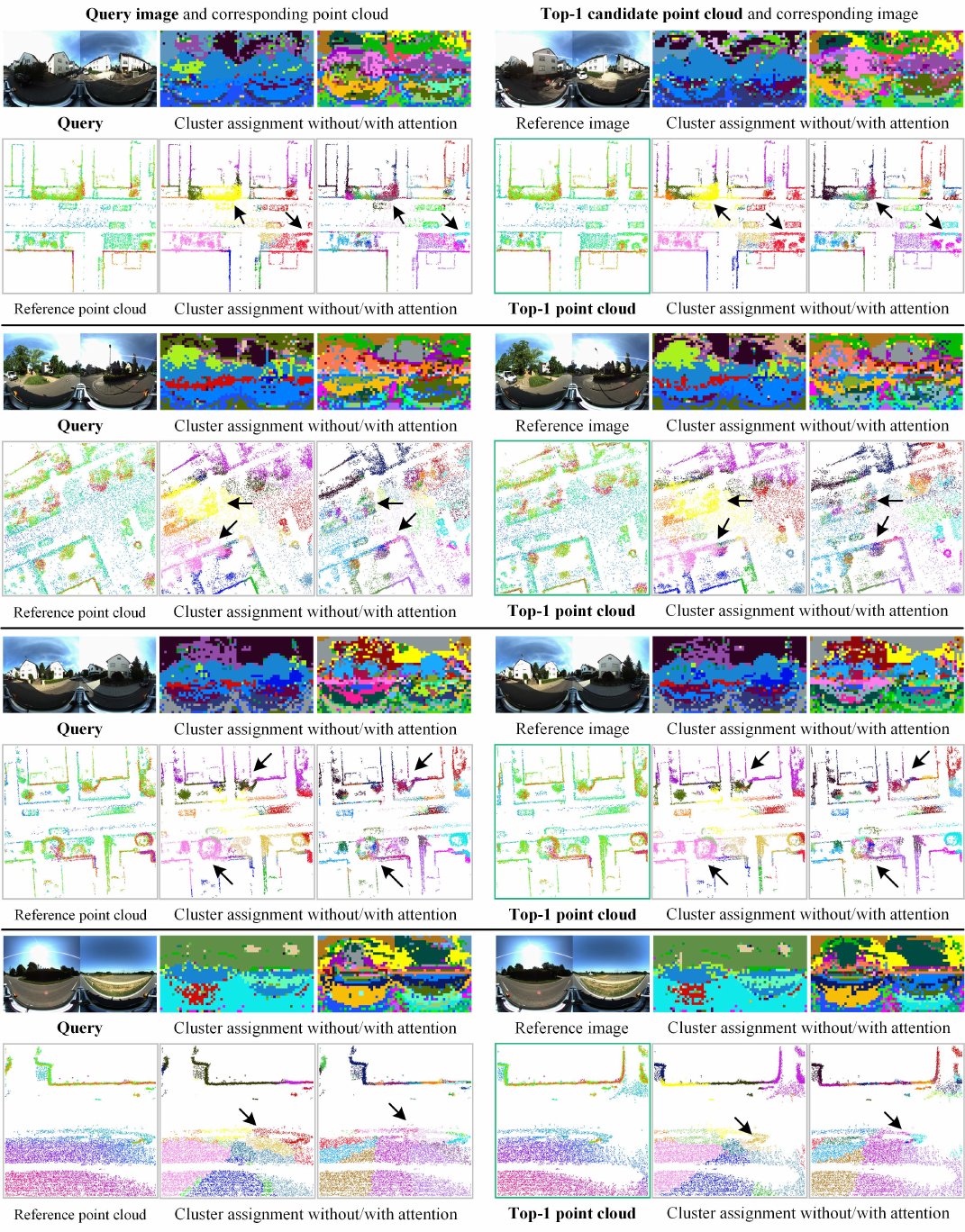}}
\caption {The VLAD cluster assignment of the query images and the Top-1 point cloud from the database. The auxiliary point clouds/images are listed for better visualization. The reference and Top-1 point clouds are rendered by the relative height, and whole point clouds are viewed from the bird eye view. The same color of patches in the cluster assignment subplots indicates the same cluster assigned.}
\label {fig_cluster_center}
\vspace{-0.5cm}
\end{figure*}

We also visualize the cluster assignment of the VLAD core generated during the local feature aggregation module in Fig.~\ref{fig_cluster_center}. The displayed 64 colors in each cluster assignment subplot represent 64 clusters, where image patches/point cloud volumes of the same color indicate the same cluster assignment. {When using the vanilla version of NetVLAD for feature aggregation, there is a tendency to aggregate multiple objects into the same cluster. However, after applying the saliency-guided feature aggregation module proposed in this paper, the clustering results more closely align with the geometric and instance distribution of objects within the scene. The arrows in the figure emphasize the areas where differences between the two methods.} The figure effectively demonstrates that our proposed saliency-guided local feature aggregation module can successfully cluster salient objects within the scene. It is worth noting that we did not explicitly incorporate semantic labels. The cluster assignment is consistent across reference and Top-1 data, which justifies the performance of our method.

\subsubsection{The efficiency of feature consistency on multi-manifold space}
{We conducted ablation experiments on different types of manifold spaces, applying feature relation consistency constraints to both Euclidean and Hyperbolic spaces.} And the results are listed in the Tab.~\ref{tab_recall_ablation}. After introducing the feature relation consistency constraints, the Recall@1 values are increased indicating that the fine-grained cross-modality feature representations are generated and are well-mapped in the high-dimensional space. The higher Recall@1 reflects the higher precision of global localization and can better satisfy the practical localization.

\subsubsection{The spherical-CNN based image tokenizer}
The baseline model~\citep{zhaoAttentionEnhancedCrossmodalLocalization2023}, employs spherical convolution to address distortion issues in panoramic images. We also designed a hybrid transformer~\citep{dosovitskiyImageWorth16x162023} architecture based on spherical convolution to extract image local patch features. Specifically, we replaced the CNN-based tokenizer in the hybrid transformer with a Spherical CNN. As shown in the Tab.~\ref{tab_recall_ablation}, leveraging the feature interaction capabilities of the Transformer minimizes the impact of feature distortion in panoramic images on cross-modality global localization tasks. However, introducing spherical convolution significantly increases the computational load of the model. Therefore, we partitioned the image directly into patches and encoded them into tokens.

\subsubsection{The number of 3D tokens}
The size of the input image is fixed and the number of patches is fixed as 2048 with default $16 \times 16$ patch size. We explore the influence of the number of sampled points. We set the number of tokens in the point cloud feature encoder as 512, 1024, 2048, 3072, and 4096 respectively. The parameter setting of the self-attention blocks are set identically. The recall performance is almost saturated at 3072 points while the training time increases significantly. 

\begin{table*}[!t]
\centering
\renewcommand{\arraystretch}{1.3}	
\caption {Ablation study of SaliencyI2PLoc that evaluated on KITTI-360 urban scenario dataset. We highlight the \textbf{best} results.}
\label{tab_recall_ablation}
\begin{threeparttable}
\resizebox{0.7\textwidth}{!}{
\begin{tabular}{lllllllll}
\toprule
SphereCNN\tnote{1}   & Attn\tnote{2}     & Euc\tnote{3}      & Hyp\tnote{4}       &R1$(\%)\uparrow$        & R5$(\%)\uparrow$  & R20 $(\%)\uparrow$    & Max $F_1$  $\uparrow$ \\\midrule 
                     &          &          &           & 64.16                  & 76.51             & 92.47                 & 0.78                  \\ \cline{1-8}
 $\surd$             &          &          &           & 60.84                  & 78.31             & 92.77                 & 0.75                  \\ \cline{1-8}
                     & $\surd$  &          &           & 66.27                  & 78.92             & 93.07                 & 0.80                  \\ \cline{1-8}         
                     &          & $\surd$  &           & 65.96                  & 80.42             & \textbf{96.08}        & 0.79                  \\ \cline{1-8}
                     &          &          & $\surd$   & 64.16                  & 77.71             & 92.17                 & 0.79                  \\ \cline{1-8}     
                     & $\surd$  & $\surd$  &           & 68.98                  & 81.02             & 93.67                 & 0.81                  \\ \cline{1-8}
                     & $\surd$  &          & $\surd$   & 68.07                  & 79.52             & 94.58                 & 0.81                  \\ \cline{1-8}
                     & $\surd$  & $\surd$  & $\surd$   & \textbf{69.28}         & \textbf{81.63}    & 93.98                 & \textbf{0.82}         \\ \bottomrule
\end{tabular}
}
\begin{tablenotes}
\smallskip\footnotesize
\item[1] SphereCNN represents using the spherical convolution-based image tokenizer.
\item[2] Attn represents adding the saliency map into the netvlad.
\item[3] Euc represents adding the relation consistency constraints on Euclidean space.
\item[4] Hyp represents adding the relation consistency constraints on Hyperbolic space.
\vspace{-0.4cm}
\end{tablenotes}
\end{threeparttable}
\end{table*}

\begin{table*}[t]
\centering
\renewcommand{\arraystretch}{1.3}	
\caption {Ablation study of SaliencyI2PLoc of a different number of point cloud tokens on KITTI-360 evaluation datasets. We highlight the \textbf{best} results.}
\label{tab_recall_ablation_token}
\begin{threeparttable}
\resizebox{0.7\textwidth}{!}{
\begin{tabular}{lllllllllll}
\toprule
 Attn\tnote{1}  & 512      & 1024      & 2048      & 3072      & 4096      & R1$(\%)\uparrow$  & R5$(\%)\uparrow$  & R{20} $(\%)\uparrow$ & Max $F_1$ $\uparrow$   \\  \midrule
                & $\surd$  &           &           &           &           & 54.52             & 74.10             & 93.37                 & 0.71                   \\
$\surd$         & $\surd$  &           &           &           &           & 63.25             & 78.61             & 92.47                 & 0.77                   \\  \hline
                &          & $\surd$   &           &           &           & 56.93             & 77.41             & 90.66                 & 0.73                   \\   
$\surd$         &          & $\surd$   &           &           &           & 64.46             & 79.22             & 93.98                 & 0.78                   \\  \hline 
                &          &           & $\surd$   &           &           & 62.95             & \textbf{80.12}    & \textbf{95.18}        & 0.77                   \\  
$\surd$         &          &           & $\surd$   &           &           & 64.76             & 79.22             & 92.47                 & 0.79                   \\  \hline 
                &          &           &           & $\surd$   &           & 64.16             & 76.51             & 92.47                 & 0.78                   \\  
$\surd$         &          &           &           & $\surd$   &           & \textbf{66.27}    & 78.92             & 93.07                 & \textbf{0.80}          \\  \hline 
                &          &           &           &           & $\surd$   & 58.43             & 69.88             & 92.47                 & 0.73                   \\   
$\surd$         &          &           &           &           & $\surd$   & 56.93             & 73.19             & 89.16                 & 0.74                   \\  \bottomrule 
\end{tabular}}
\begin{tablenotes}
\footnotesize
\item[1] Attn represents adding the saliency map into the netvlad.
\end{tablenotes}
\end{threeparttable}
\vspace{-0.4cm}
\end{table*}

\begin{table}[!t]
\centering
\renewcommand{\arraystretch}{1.3}	
\caption {The parameters of the model and inference time.}
\label{tab_time_analysis}
\resizebox{0.48\textwidth}{!}{
\begin{tabular}{lllllllll}
\toprule
\multirow{2}{*}{Method}     & \multicolumn{3}{c}{Params (M)}        &  \multirow{2}{*}{Inference FPS$\uparrow$}         \\ \cline{2-4}
                            & Visual & Point cloud  & Total         &                                                   \\ \midrule 
AE-Spherical ~\citep{zhaoAttentionEnhancedCrossmodalLocalization2023}       & 19.76     & 20.04     &   39.80   & 84.49  \\ 
\textbf{SaliencyI2PLoc} (ours)                                             & 14.58     & 28.95     &   43.54   & 24.15  \\ \bottomrule
\end{tabular}%
}
\vspace{-0.4cm}
\end{table}

\begin{figure}[!t]
\centering
\includegraphics[scale=0.45]{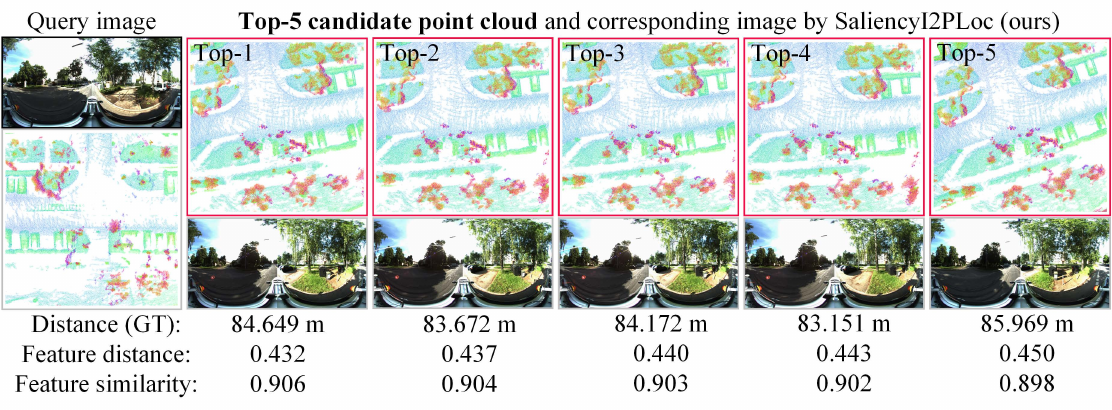}
\vspace{-0.4cm}
\caption {The failure case of our method.}
\label {fig_failure_case}
\vspace{-0.5cm}
\end{figure}

\subsection{Limitations}
The Tab.~\ref{tab_time_analysis} lists the parameters of different modules in SaliencyI2PLoc. The parameters of our model are comparable to those of the baseline model. With a full 24GB GPU memory usage, the baseline model achieves approximately 84.49 FPS, whereas our method achieves 24.15 FPS. The increased inference time is due to the use of a self-attention mechanism to capture long-range features. The implementation of the self-attention block could be optimized by employing the recently proposed flash attention~\citep{daoFlashAttentionFastMemoryefficient2022}, and the speed of our model can be further improved.

We also present the failure case of our method, as shown in Fig.~\ref{fig_failure_case}. The two scenarios are very similar, both situated at an intersection, with trees and roads in the same direction and a very similar overall layout, making visual differentiation quite difficult. As a result, in these scenarios, both our method and the baseline approach fail. Post-processing strategies, such as re-rank mechanisms, may solve these problems.

\section{Conclusion}
\label{sec_con}
This paper follows classical two-tower architecture and proposes a novel cross-modality fusion and image-to-point cloud {coarse} global localization method, SaliencyI2PLoc, by leveraging the contrastive learning feature representation framework. The context saliency map is fused into local patch feature aggregation to focus on the objects that contribute more to global scene representation. Besides, the feature relation consistency between image-point cloud pairs of different modalities in multi-manifolds is considered to facilitate feature fusion and alignment. Experiments on datasets with different types of scenes have shown the effectiveness and robustness of this method. In the future, we will attempt to introduce knowledge distillation to squeeze the model size to facilitate the usage of mobile robots. Besides, we will explore fusing the 2D-3D correspondence matching module to achieve pose estimation. 


\printcredits

\section{Declaration of interests}
The authors declare that they have no known competing financial interests or personal relationships that could have appeared to influence the work reported in this paper.

\bibliographystyle{cas-model2-names}

\bibliography{ref}

\end{sloppypar}
\end{document}